\documentclass[lettersize,journal]{IEEEtran}
\usepackage{amsmath,amsfonts}
\usepackage{algorithmicx}
\usepackage{algorithm}
\usepackage{algpseudocode}

\usepackage{array}
\usepackage[caption=false,font=scriptsize,labelfont=sf,textfont=sf]{subfig}
\usepackage{textcomp}
\usepackage{stfloats}
\usepackage{url}
\usepackage{verbatim}
\usepackage{graphicx}
\usepackage{cite}
\usepackage[utf8]{inputenc}
\usepackage{ragged2e}
\usepackage{microtype}
\usepackage{pgfplots}
\usepackage{comment}
\usepackage{xcolor}
\usepackage{colortbl}
\definecolor{gray}{rgb}{0.9,0.9,0.9}

\usepackage{adjustbox}
\usepackage{amssymb}
\usepackage{amsmath}
\usepackage{multirow}

\begin{document}

\title{Semi-Supervised Domain Adaptation Using Target-Oriented Domain Augmentation for 3D Object Detection}

\author{Yecheol Kim$^{1*}$, Junho Lee$^{1*}$, Changsoo Park$^2$, Hyoung won Kim$^2$, Inho Lim$^2$, Christopher Chang$^2$, \\ and Jun Won Choi$^3$ \\
\thanks{$^1$Y. Kim and J. Lee are with Department of Electrical Engineering, Hanyang University, 04753 Seoul, Republic of Korea. (e-mail: yckim@spa.hanyang.ac.kr, jhlee@spa.hanyang.ac.kr)}

\thanks{$^2$C. Park, H. Kim, I. Lim and C. Chang are with Kakao Mobility Corp, Pangyo 13529, Republic of Korea. (e-mail: teddy.p@kakaomobility.com, gemini.k@kakaomobility.com, ed.lim@kakaomobility.com, cswchang@alumni.caltech.edu)}

\thanks{$^3$J. W. Choi is with Department of Electrical and Computer Engineering, College of Liberal Studies, Seoul National University, Seoul, 08826, Korea. (e-mail: junwchoi@snu.ac.kr)\textit{(Corresponding author: Jun Won Choi)}}

\thanks{$^*$denotes equal contribution.}
}

\markboth{Journal of \LaTeX\ Class Files,~Vol.~14, No.~8, April~2024}%
{Shell \MakeLowercase{\textit{et al.}}: A Sample Article Using IEEEtran.cls for IEEE Journals}

\IEEEpubid{0000--0000/00\$00.00~\copyright~2024 IEEE}

\maketitle

\begin{abstract}

3D object detection is crucial for applications like autonomous driving and robotics. However, in real-world environments, variations in sensor data distribution due to sensor upgrades, weather changes, and geographic differences can adversely affect detection performance. Semi-Supervised Domain Adaptation (SSDA) aims to mitigate these challenges by transferring knowledge from a source domain, abundant in labeled data, to a target domain where labels are scarce. This paper presents a new SSDA method referred to as Target-Oriented Domain Augmentation (TODA) specifically tailored for LiDAR-based 3D object detection. TODA efficiently utilizes all available data, including labeled data in the source domain, and both labeled data and unlabeled data in the target domain to enhance domain adaptation performance. TODA consists of two stages: TargetMix and AdvMix. TargetMix employs mixing augmentation accounting for LiDAR sensor characteristics to facilitate feature alignment between the source-domain and target-domain. AdvMix applies point-wise adversarial augmentation with mixing augmentation, which perturbs the unlabeled data to align the features within both labeled and unlabeled data in the target domain. Our experiments conducted on the challenging domain adaptation tasks demonstrate that TODA outperforms existing domain adaptation techniques designed for 3D object detection by significant margins. The code
is available at: https://github.com/rasd3/TODA.

\end{abstract}


\section{Introduction}
\IEEEPARstart{3}{D} object detection is the task of detecting and localizing objects in 3D world coordinates using sensor measurements. 3D object detection has risen as a pivotal perception task in the field of autonomous vehicles and robotics.   Recently, 3D point cloud data acquired by LiDAR sensor has been successfully used to achieve promising performance in 3D object detection. The recent progress of deep learning has sparked the development of a plethora of architectures for detecting objects from LiDAR point cloud. Widely used 3D object detectors include VoxelNet\cite{voxelnet}, PointPillar \cite{pointpillars}, SECOND \cite{second}, CenterPoint \cite{centerpoint}, PV-RCNN \cite{pvrcnn}, PillarNet \cite{pillarnet}, and Voxel R-CNN \cite{voxelrcnn}.

\begin{figure}[t]
  \begin{center}
    \includegraphics[width=0.98\columnwidth]{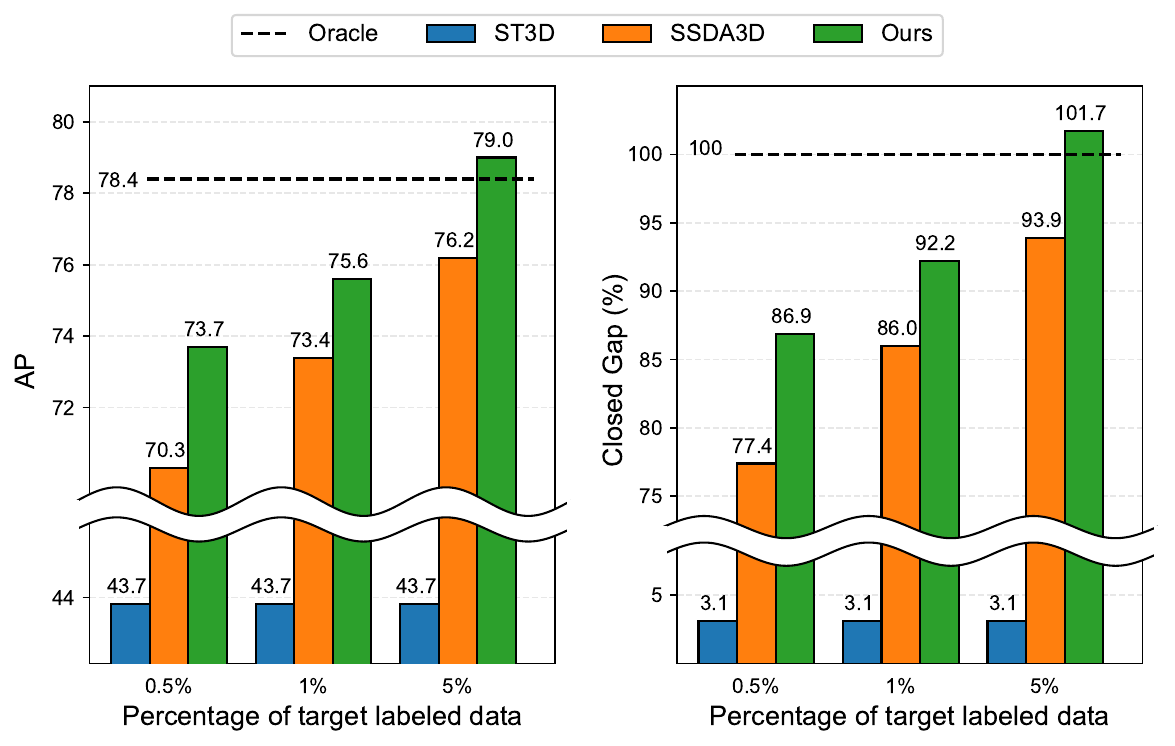}
  \end{center}
    \caption{{\bf Performance evaluation in a domain adaptation task from Waymo dataset to nuScenes dataset:} 0.5\%, 1\%, and 5\% labeled data in the target domain are used.  A SSDA method using only 0.5\% of the target label results in a remarkable performance gain over a UDA method (ST3D \cite{st3d}). Our TODA also significantly outperforms SSDA3D \cite{ssda3d} in all settings. Surprisingly, TODA even surpasses the \textit{Oracle} performance with only 5\% labels.}
    \label{fig:bar_plot}
\end{figure}

\IEEEpubidadjcol

A shift in the distribution of data often leads to notable decreases in the performance of 3D object detection \cite{shift1,sn}. In the context of autonomous driving, shifts in distribution arise from change in sensor suites, fluctuations in weather conditions, disparities in geographical locations, and more. For instance, upgrades in sensor specifications, including resolution, field of view (FOV), and intensity, introduce shifts in data distribution. In this case, it is inefficient to collect new training data and retrain the model from scratch with each sensor replacement.  Therefore, addressing the domain shift problem in 3D object detection is crucial for the commercial deployment of autonomous driving and still remains a significant open challenge.



Domain adaptation offers a solution to this problem by allowing models trained in source domains to adapt effectively to different but related target domains, thereby reducing the need for extensive data labeling. Domain adaptation strategies can be categorized into two main types: {\it unsupervised domain adaptation} (UDA) and {\it semi-supervised domain adaptation} (SSDA).

When labeled data is unavailable within the target domain, UDA transfers knowledge learned from labeled source domains  to enhance performance in target domains. Recent UDA approaches for 3D object detection include CL3D \cite{cl3d}, ST3D \cite{st3d}, and ST3D++ \cite{st3d++}. These methods have employed various domain-adaptive pseudo-labeling approaches to mitigate domain discrepancies.

While UDA can mitigate domain shift issues, addressing significant domain gaps between source and target domains  remains challenging. While UDA is effective in narrowing the performance gap from the \textit{Oracle}\footnote{The \textit{Oracle} model denotes the fully-supervised model trained on the target domain.} by approximately 80\% when applied to similar LiDAR specifications from the Waymo dataset \cite{waymo} to the KITTI dataset \cite{kitti}, it achieves only a 3\% reduction in the gap when dealing with markedly different LiDAR configurations, such as those from the Waymo dataset to the nuScenes dataset \cite{nuscenes}.
To address this limitation, semi-supervised domain adaptation (SSDA) has emerged as a cost-effective way to improve the effect of domain adaptation.  Unlike UDA, SSDA uses a small amount of labeled target-domain data along with a substantial volume of unlabeled target-domain data to improve performance of domain adaptation. Fig. \ref{fig:bar_plot} illustrates that leveraging a small amount of labeled target-domain data, SSDA methods can yield substantial performance improvements over UDA. 

SSDA3D \cite{ssda3d}, as of now, is the sole  existing SSDA method specifically designed for 3D object detection. This method operates in a two-stage process. Initially, it incorporates an {\it Inter-domain Point-Cutmix} operation to reduce domain bias and thus learns domain-invariant representations.  Following this, SSDA3D employs an {\it Intra-domain Point-MixUp} operation, which combines both labeled data and pseudo-labeled scenes on a global scale under a semi-supervised learning (SSL) framework.
Despite its notable performance gains over UDA methods, we claim that SSDA3D has not fully exploited the distinct properties inherent in LiDAR point cloud data.

In this study, we present a novel SSDA framework referred to as {\it Target-Oriented Domain Augmentation} (TODA) for 3D object detection.  TODA employs a two-stage data augmentation strategy for SSDA;  \textit{Target Sensor-Guided Mix Augmentation} (TargetMix) and  \textit{Adversarial-Guided Mix Augmentation} (AdvMix).

TargetMix reduces the disparity between the source and target domains by employing a cross-domain mixup strategy. Initially, TargetMix aligns the characteristics of LiDAR point clouds, such as Field of View (FOV) and beam configurations in the source-domain data with those in the target-domain data. Subsequently, a cross-domain mixup augmentation is applied in a polar coordinate system, incorporating an effective LiDAR distribution matching process that considers the scanning mechanism of LiDAR technology. By generating convex combinations of LiDAR point clouds from both source and target domains, TargetMix ensures smooth transitions between these domains. 


TargetMix does not utilize the unlabeled data in the target domain, leaving room for further improvement. AdvMix employs a pseudo-labeling approach to leverage this unlabeled data. However, as pointed out in \cite{ape}, the pseudo-labeling approach might suffer from intra-domain discrepancy, which arises when the teacher model, trained solely with labeled data, yields feature points divided into those attracted into the source domain and those that do not. Such inconsistency within the target domain data diminishes the effectiveness of pseudo labeling. To address this issue, we introduce AdvMix, a technique that integrates adversarial point augmentation with mixup augmentation. {\it Adversarial Point Augmentation}, derived from {\it Adversarial Augmentation} in \cite{biat}, involves perturbing unlabeled data at the point level. These perturbations are informed by the negative gradient of the detection loss calculated in terms of the unlabeled points, effectively altering their distribution to enhance detection performance. This process generates perturbed pseudo-labeled samples that help minimize intra-domain discrepancies. Following this, we implement mixup augmentation, blending the labeled data with the pseudo-labeled data on a global scale.



We evaluate the performance of TODA on the challenging domain adapation tasks: from the Waymo dataset \cite{waymo} to the nuScenes dataset \cite{nuscenes} and from the nuScenes dataset to the KITTI dataset \cite{kitti}. Our results demonstrate that TODA yields substantial performance improvements compared to the baseline method. Moreover, TODA outperforms existing domain adaptation techniques, including SSDA3D, by considerable margins.

The key contributions of TODA are summarized as follows:
\begin{itemize}
    \item We propose a novel two-stage SSDA framework for 3D object detection.
    \item We present two novel data augmentation techniques, TargetMix and AdvMix, designed to improve the effectiveness of domain adaptation. TargetMix aims to reduce the domain disparity by leveraging labeled data from both the source and target domains. In contrast, AdvMix uses a combination of labeled and unlabeled data within the target domain to minimize the intra-domain gap.
    \item     TargetMix method is specifically designed to take into account the unique  characteristics of LiDAR point cloud data, thereby maximizing the impact of data augmentation. 
    \item AdvMix generates adversarial examples to align the representation of unlabeled data effectively within the target domain. In our study, we are the first to pioneer the use of adversarial augmentation for point cloud-based SSDA.
    \item The proposed TODA achieves state-of-the-art performance on popular domain adaptation benchmarks. It attains performances on par with the {\it Oracle} performance even when utilizing merely 5\% of labeled data in the target domain.
\end{itemize}

\section{Related Work}

\subsection{LiDAR-based 3D Object Detection}

The advancement of deep learning has led to the development of various LiDAR-based 3D object detectors. These detectors utilize point encoding to extract high-level semantic features. Point encoding techniques can generally be categorized into two types: clustering-based and grid-based methods. Clustering-based methods group point clouds into clusters and encode points within each cluster in a hierarchical manner. Examples of clustering-based methods include PointRCNN \cite{pointrcnn}, 3DDS \cite{3dssd}, and PV-RCNN \cite{pvrcnn}. In contrast, grid-based encoding methods divide 3D point clouds into voxel or pillar grids, thererby producing features in a structured grid pattern. 3D object detectors that use grid-based encoding include VoxelNet \cite{voxelnet}, SECOND \cite{second}, PointPillar \cite{pointpillars}, and CenterPoint \cite{centerpoint}.


\begin{figure*}[t]
  \begin{center}
    \includegraphics[width=0.95\textwidth]{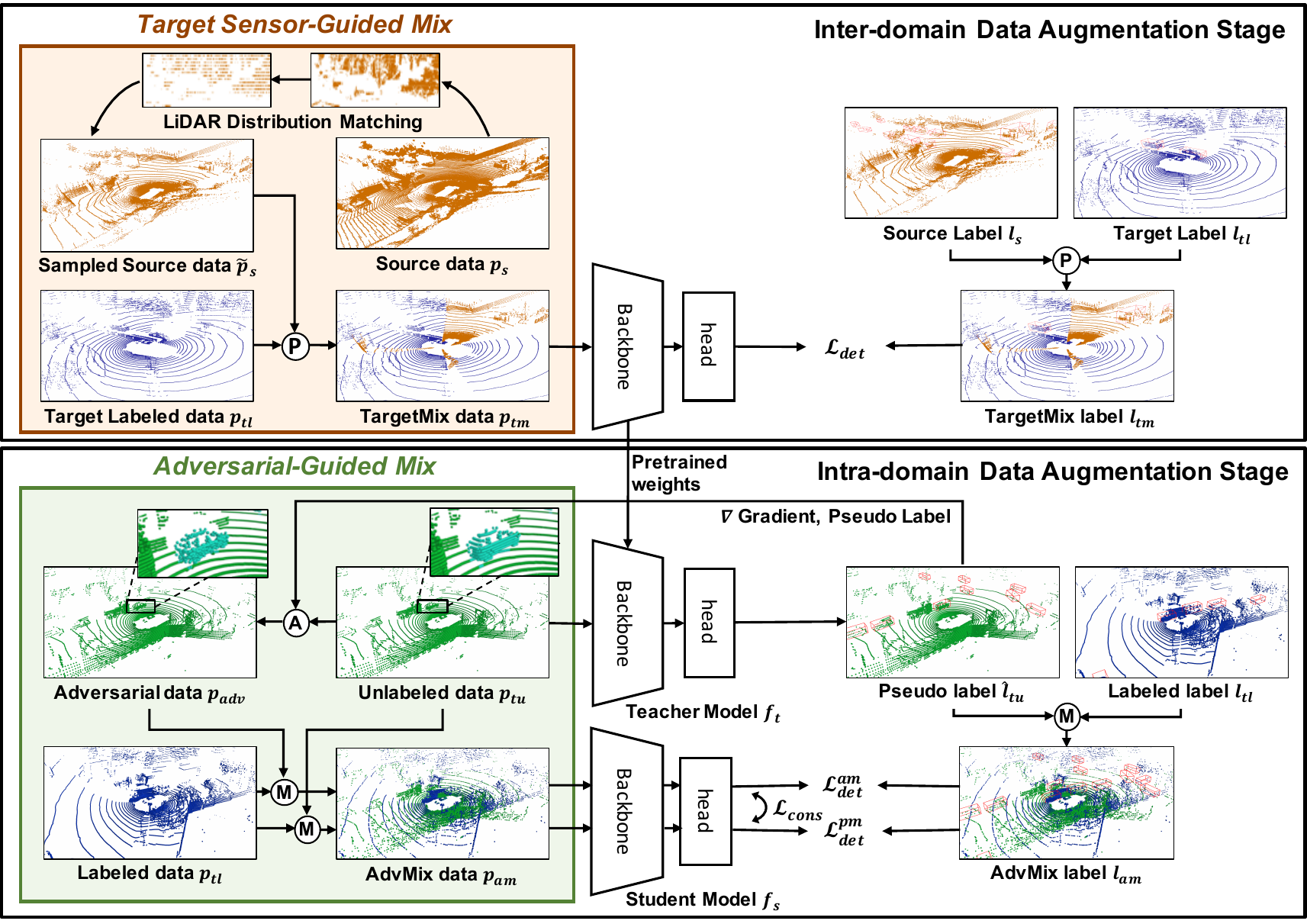}
  \end{center}
    \caption{{\bf Overall architecture of the proposed TODA:}  First, TargetMix aligns the source-domain data with target-domain data by applying LiDAR Distribution Matching, followed by mixup augmentation in polar coordinates. Then, AdvMix utilizes Adversarial Point Augmentation to perturb the unlabeled data in the target domain, aiming to produce consistent representation of both labeled and unlabeled data.  'P', 'A', and 'M' denote \textit{Polar Coordinate-based Mix}, \textit{Adversarial Point Augmentation}, and \textit{Point-Mixup} respectively.}
    \label{fig:framework}
\end{figure*}

\subsection{Domain Adaptation for 3D Object Detection}
Several UDA methods have proposed for 3D object detection. PointDAN \cite{pointdan} and SRDAN \cite{srdan} transformed object features taking factors like the range and distance of point clouds into account. ST3D \cite{st3d} and ST3D++ \cite{st3d++} tackled the challenge of learning instability induced by domain shift through the utilization of a memory bank and denoising techniques. Currently, only one SSDA method available in the literature is SSDA3D \cite{ssda3d}. SSDA3D employed mixing augmentation strategies to reduce the distribution gap between the source and target domains.

\subsection{Mixing Augmentation for Domain Adaptation}


Mixing augmentation has emerged as an effective strategy for addressing data limitations and enhancing robustness in semi-supervised learning (SSL) and domain adaptation. Mixup \cite{mixup} generated new training data by creating convex combinations of input pairs, subsequently training the model on these blended inputs and their corresponding labels. CutMix \cite{cutmix}, a variant of Mixup for image recognition, constructed new images by randomly cutting a rectangular patch from one image and pasting it onto another at a random location. These methods encouraged the model to learn interpolated decision boundaries between samples and hence improved generalization performance significantly.

In domain adaptation, mix augmentation techniques are employed to bridge inter-domain gaps by integrating data from both the source and target domains, thereby facilitating knowledge transfer from the source to the target domain. PolarMix \cite{polarmix} introduces a specialized mix augmentation strategy for point cloud data that combines scene-level and object-level features in cylindrical coordinates.



\subsection{Adversarial Augmentation}

Adversarial augmentation is a method for generating \textit{adversarial examples} that are intended to challenge the  accuracy of a model. 
Recently, AT \cite{at} and VAT \cite{vat} have utilized a gradient-based perturbation generation method to enhance the robustness of models in both supervised and semi-supervised tasks.
Furthermore, several studies \cite{eaad, asp, apc, advpc} have explored using adversarial augmentation strategies to generate and translate point clouds.



\subsection{Semi-Supervised Learning for 3D Object Detection}
SSL has also been actively studied for 3D object detection. SSL involves using a combination of a small amount of labeled data and a large amount of unlabeled data for training models. SESS \cite{sess} introduced a \textit{Consistency Loss} as a means to align predicted 3D object proposals between the teacher and student networks. 3DIoUMatch \cite{3dioumatch} filtered low-quality pseudo-labels using the combination of confidence thresholds and 3D IoU predictions. Proficient Teachers \cite{proteachers} enhanced the precision of pseudo-labels through an augmented prediction approach that incorporates box voting-based ensembling.

\section{Method}
In this section, we present the details of the proposed TODA method.

\subsection{Overview}
In the SSDA framework, we train the model using labeled data from the source domain, as well as both labeled and unlabeled data from the target domain. First, the $N_S$ labeled point cloud samples in the source domain are denoted as $D_{S}=\{p^i_{s},l^i_{s}\}^{N_S}_{i=1}$, where $p^i_{s}$ and $l^i_{s}$ are the $i$th point clouds and the corresponding 3D object detection labels. 
Similarly, the $N_{TL}$ labeled samples in the target domain are denoted as $D_{TL}=\{p_{tl}^i,l_{tl}^i\}^{N_{TL}}_{i=1}$  and the $N_{TU}$ unlabeled samples in the target domain are denoted as $D_{TU}=\{p_{tu}^i\}^{N_{TU}}_{i=1}$. The $i$th point clouds $p^i\in\mathbb{R}^{N_{p}\times4}$ contain LiDAR points where each point measurement includes  3D coordinates $(x, y, z)$, and intensity $I$. The corresponding set of labels, $l^i$ contains descriptions of 3D object boxes, consisting of category, location, size, and heading angle. In typical SSDA setup, we assume that both $N_{S}$ and $N_{TU}$ are significantly larger than $N_{TL}$, i.e. $N_S>>N_{TL}$ and $ N_{TU}>>N_{TL}$. The objective of SSDA is to maximize the object detection performance in the target domain through an effective use of $D_{S}$, $D_{TL}$,  and $D_{TU}$. 

Fig. \ref{fig:framework} depicts the two-stage structure of the proposed TODA framework.
In the first stage, TargetMix initially performs LiDAR Distribution Matching, which transforms the source-domain LiDAR data $D_{S}$ into the LiDAR data $D'_{S}$ such that  the transformed LiDAR data $D'_{S}$ follows the configurations of the target domain LiDAR. Subsequently, TargetMix combines the transformed data $D'_{S}$ with $D_{TL}$  in polar coordinates using a weighted mixup operation as proposed in \cite{cutmix}. This operation results in a mixed dataset $D_{TM}=\{p_{tm}^i, l_{tm}^i\}^{N_{TM}}_{i=1}$.  The mixed data $D_{TM}$ is then utilized to train a 3D object detection model, aimed at bridging the gap between the two domains.

In the second stage, we utilize two 3D object detection pipelines with identical structures. As illustrated in Fig. \ref{fig:framework}, one model  $f_t$ serves as a teacher, while the other $f_s$ functions as a student model. The weights of the model trained in the first stage are initially copied to the teacher model. The teacher model is then employed to generate pseudo-labels  $\{\hat{l}^i_{tu}\}^{N_{TU}}_{i=1}$ from the unlabeled data $D_{TU}$. At the same time, the corresponding unlabeled data $D_{TU}$ are perturbed by the Adversarial Point Augmentation. These perturbed samples are then mixed with the labeled data $D_{TL}$, resulting in the mixed data $D_{AM}=\{p_{am}^i,\hat{l}_{tu}^i\}^{N_{TU}}_{i=1}$.
Finally, the student model is trained using the  mixed data $D_{AM}=\{p_{am}^i,\hat{l}_{tu}^i\}^{N_{TU}}_{i=1}$. 
By incorporating adversarial examples, TODA ensures a consistent representation of the unlabeled samples to train the student model, facilitating the effective utilization of unlabeled target-domain samples.


\begin{figure}[t]
    \centering
    \includegraphics[width=0.9\columnwidth]{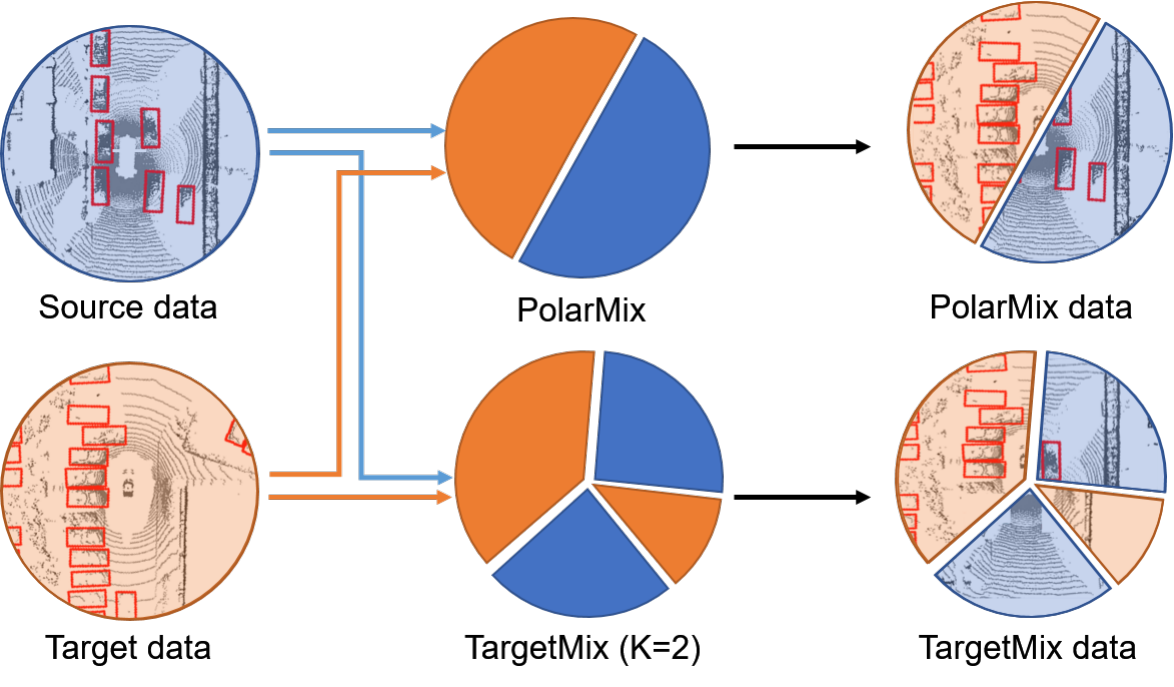}
    \caption{{\bf Comparison of TargetMix with PolarMix:} TargetMix divides the entire azimuth angle into $2K$ separate sectors while PolarMix divides it into two sectors.}
    \label{fig:targetmix_sep}
\end{figure}

\subsection{Target Sensor-Guided Mix} 
TargetMix combines LiDAR points from source and target domains. 
TargetMix performs two steps. The first step converts the source-domain LiDAR data $D_{S}$  into the data $D'_{S}=\{\tilde{p}^i_s, l^i_s\}^{N_S}_{i=1}$. This LiDAR Distribution Matching step adjusts the number of beam channels, the number of points per channel, and Vertical Field of View (VFOV) of $D_{S}$ to match the corresponding parameters of the LiDAR data in the target domain.
Specifically, 
TargetMix converts the 3D points in $D_S$ from Cartesian coordinates $(x,y,z)$ to spherical coordinates:
\begin{align}
& \theta = \arctan \frac{z}{\sqrt{x^2 + y^2}}, \quad \phi = \arcsin \frac{y}{\sqrt{x^2 + y^2}},
\end{align}
where $\theta$ and $\phi$ denote the azimuth and zenith angles, respectively. Using the point data in spherical coordinates, we then generate the range image $I_r \in \mathbb{R}^{H \times W}$, where each pixel in $I_r$ corresponds to specific $\theta$ and $\phi$ values. The image $I_r$ is downsampled based on the ratio of the number of beam channels, the number of points per channel, and the VFOV. 
The downsampled image $I'_r$ is transformed back into the Cartesian coordinate system to  generate the transformed source data $D'_{S}$. For instance, consider the Waymo dataset as source-domain data and the nuScenes dataset as target-domain data. LiDAR sensor used in Waymo dataset has 64 channels, about 2200 points per channel, and a VFOV of $[-17.6^\circ, 2.4^\circ]$, covering a range of $20^\circ$. On the other hand, LiDAR sensor in nuScenes dataset has 32 channels, 1100 points per channel, and a VFOV of $[-30.0^\circ, 10.0^\circ]$, covering a range of $40^\circ$. 
The image $I_r$ is downsized vertically by a factor of $4$, considering both the VFOV range ratio $\frac{40^\circ}{20^\circ}$ and the channel ratio $\frac{64}{32}$. Additionally, the image is downsized horizontally by a factor of $2$ to match the points per channel between the two datasets. 

The second step of TargetMix applies mix augmentation between $D'_{S}$ and $D_{TL}$ in the polar coordinate system, where each LiDAR point is characterized by $(\theta,r,\phi)$: $\theta$ represents the azimuth angle, $r$ denotes the distance, and $\phi$ indicates the inclination angle between the z-axis and the point vector $(x,y,z)$. Inspired by PolarMix \cite{polarmix}, TargetMix conducts a mix operation by partitioning the azimuth angle into two sections and filling one with data points from $D'_{S}$ and the other with those from $D_{TL}$. This mix operation is also applied to 3D box labels accordingly.
Unlike PolarMix, which utilizes a single contiguous sector for $D_{TL}$, TargetMix assigns $K$ separate contiguous sectors to $D_{TL}$, accounting for the widespread distribution of objects. The remaining regions are assigned to $D'_{S}$. However, such partitioning can lead to ambiguity, as LiDAR points within objects may be divided by azimuth boundaries. To mitigate this issue, TargetMix not only eliminates the object boxes affected by these boundaries but also removes their corresponding point cloud data. This operation is referred to as {\it Enhanced Mix Strategy}. We denote the data generated by Enhanced Mix Strategy as  $D_{TM}=\{p_{tm}^{i}, l_{tm}^{i}\}_{i=1}^{N_{TM}}$. Note that 
\begin{gather}
 p_{tm}^{i}=(M_p \odot \tilde{p}^i_s) \oplus ((1-M_p) \odot p^j_{tl})\\
 l_{tm}^{i}=(M_l \odot l^i_s) \oplus ((1-M_l) \odot l^j_{tl})
\end{gather}
where $M_p, M_l$ denotes a binary mask list indicating the $K$ azimuth ranges $[[\alpha^1,\beta^1],...[\alpha^K,\beta^K]]$, $\odot$ is the element-wise multiplication operation, and $\oplus$ denotes the concatenation operation. TargetMix randomly selects either a mixed sample from $D_{TM}$ or a sample from $D'_{S}$ with a probability $P_{tm}$, and the chosen sample is fed into the 3D object detection model for training.




\subsection{Adversarial-Guided Mix} 
\begin{figure}[t]
        \subfloat[][]{\includegraphics[width=0.5\columnwidth]{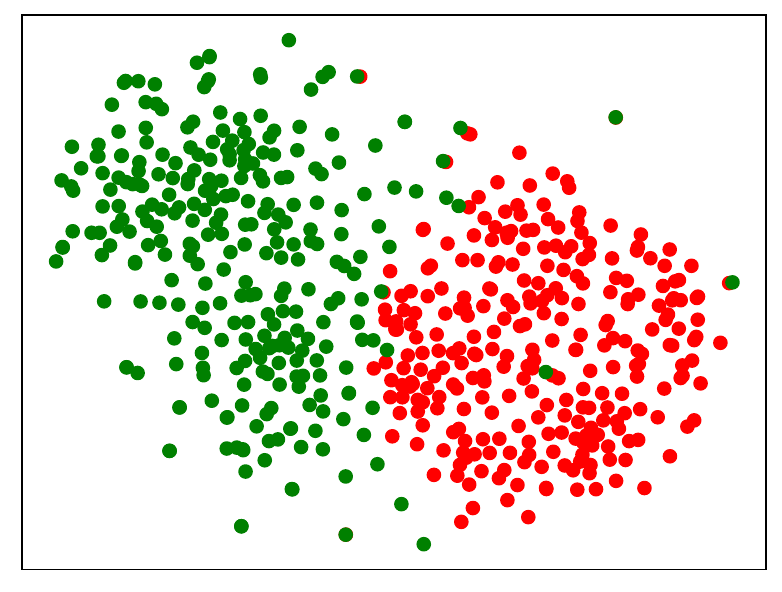}
        \label{fig:tsne_org}}
        \subfloat[][]{\includegraphics[width=0.5\columnwidth]{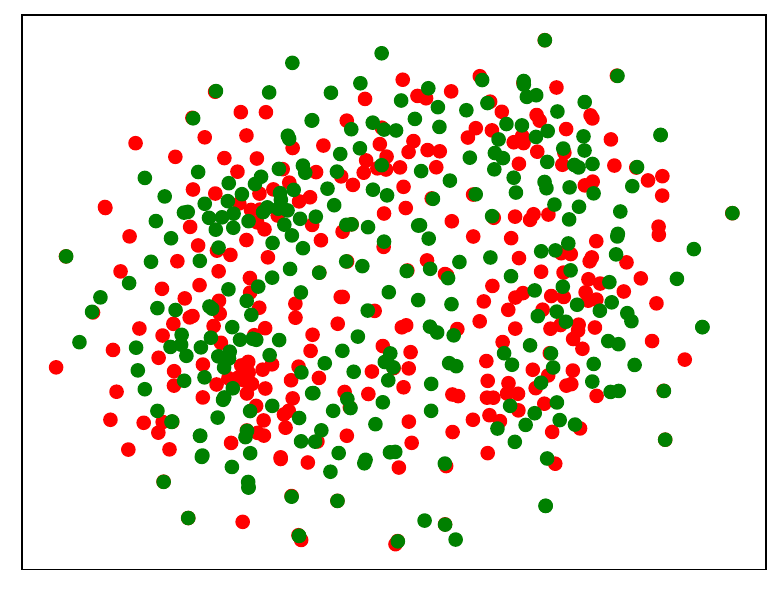}
        \label{fig:tsne_adv}}
    \caption{{\bf t-SNE visualization of features:} (a)  unlabeled data (green) versus  labeled data (red), (b) adversarial examples (green) versus labeled data (red) within the target domain. These features are extracted from the final layer of the teacher model trained with TargetMix.}
    \label{fig:tsne}
\end{figure}

The teacher model $f_t$ is initialized with the model trained by TargetMix in the first stage. Then, the teacher model $f_t$ generates pseudo labels $\{\hat{l}^i_{tu}\}^{N_{TU}}_{i=1}$ from the unlabeled target data $D_{TU}$. However, because the teacher model is trained on a limited subset of labeled data in both the source and target domains, it may not fully capture the distribution of the unlabeled target data. This results in intra-domain discrepancy, where the feature points from labeled data exhibit a different distribution from those of the unlabeled data. Since the teacher model is trained with labeled data, some unlabeled data may also be influenced by the labeled data while others are not. This discrepancy is evident in Fig. \ref{fig:tsne} (a), where a noticeable distribution gap exists between the representation of the labeled data and that of the unlabeled data. Such intra-domain inconsistency presents challenges when training the student model using the inconsistently represented labeled and unlabeled data.

To address this issue, we employ adversarial augmentation as proposed in \cite{biat}. This technique  aligns data distribution with the target distribution by perturbing input data in the direction of the negative gradient of the loss function derived from a model trained on the target distribution. AdvMix specifically implements Adversarial Point Augmentation on unlabeled LiDAR data within the target domain, altering the positions of LiDAR points in 3D space to optimize domain adaptation performance.
More precisely, LiDAR points located within detection boxes provided by pseudo labels, are subject to perturbation. Within each bounding box, a subset of LiDAR points is randomly chosen to be perturbed, based on a probability $\rho$.
Subsequently, for each selected point, one of three types of adversarial perturbations is applied
\begin{itemize}
    \item Point translation:  a perturbation $\delta$ is added to the $(x, y, z)$ coordinates of the selected point
    \item Point addition: a new point is generated by translating the coordinates of the selected point by $\delta$
    \item Point removal: the selected point is removed.
\end{itemize}
One of the three types is randomly chosen with equal probability, producing the adversarial samples $D_{adv}=\{p_{adv}^{i},l_{adv}^{i}\}_{i=1}^{N_{TU}}$
The direction of the perturbation $\delta$ is determined by the negative gradient of the detection loss.  For the point cloud input $p_{tu}$, the perturbation $\delta$ can be calculated as
\begin{align}
\delta &=\epsilon\frac{g}{||g||_{2}}, \\
g &=-\nabla_{p_{tu}}\mathcal{L}_{det}(f_{t}(p_{tu}),\hat{l}_{tu}),
\end{align}
where $\mathcal{L}_{det}$ is the detection loss comprising the classification and regression losses, $\hat{l}_{tu}$ includes the pseudo labels obtained from $p_{tu}$, $g\in\mathbb{R}^3$ is the gradient of the detection loss with respect to the $(x,y,z)$ coordinates of the points in $p_{tu}$, and $\epsilon$ denotes the magnitude of the perturbation.
This gradient indicates the direction of perturbation that reduces the detection loss.  Fig. \ref{fig:tsne} (b) shows that the Adversarial Point Augmentation results in the reduced gap in distribution between the labeled and unlabeled data. 
Consequently, this can foster a more consistent and aligned learning process for the student model.

\begin{algorithm}[t]
\caption{Target Oriented Data Augmentation (TODA)}\label{alg:cap13}
\textbf{Input:} Source data $(p_s, l_s) \in D_S$, Target labeled data $(p_{tl}, l_{tl}) \in D_{TL}$, Target unlabeled data $p_{tu} \in D_{TU}$, Teacher model $f_t$, Student model $f_s$, the probability of TargetMix $P_{tm}$, the probability of AdvMix $P_{am}$, Regularization parameter $\lambda$, total epoch of TargetMix stage $T_{\text{TM}}$, total epoch of AdvMix stage $T_{\text{AM}}$
\begin{algorithmic}[1]
\State Apply DataTransformation: $D'_s$ \Comment{TargetMix stage}
\For{$T = 1$ to $T_{\text{TM}}$} 
    \For{$i = 1$ to $N_{S}+N_{TL}$}
        \If{rand() $< P_{tm}$} 
            \State Sample $(p'^{i}_{s}, l^{i}_{s})$ and $(p^{i}_{tl}, l^{i}_{tl})$ 
            \State Apply PolarEnhanceMix: ($p'^{i}_{tm}$, $l^{i}_{tm}$)
        \Else 
            \State Sample $(p'^{i}_{s}, l^{i}_{s})$ if $i < N_{S}$ else $(p^{i}_{tl}, l^{i}_{tl})$
        \EndIf
        \State Calculate $\mathcal{L}_{det}$ and update $f_t$
    \EndFor
\EndFor
\State Initialize $f_s$ with $f_t$ pre-trained weights \Comment{AdvMix stage}
\State Generate pseudo labels $\hat{l}_{tu}$ using $f_t$
\For{$T = 1$ to $T_{\text{AM}}$} 
    \For{$i = 1$ to $N_{TL}+N_{TU}$}
        \State Sample $(p^{i}_{tl}, l^{i}_{tl})$ and $(p^{i}_{tu}, \hat{l}^{i}_{tu})$
        \State Generate adversarial example $p^{i}_{adv}$
        \If{rand() $< P_{am}$} 
            \State Apply PointMixUp: ($\Tilde{p}^{i}_{AM}$, $\hat{l}^{i}_{AM}$), ($\Tilde{p}^{i}_{PM}$, $\hat{l}^{i}_{PM}$)
            \State Calculate $\mathcal{L}_{cons}(f_s(\Tilde{p}^{i}_{AM}),f_s(\Tilde{p}^{i}_{PM}))$
        \Else
            \State Calculate $\mathcal{L}_{cons}(f_s(p^{i}_{tu}),f_s(p^{i}_{adv}))$
        \EndIf
        \State Calculate $\mathcal{L}_{det}$ and update $f_s$
    \EndFor
\EndFor
\end{algorithmic}
\end{algorithm}

Finally, we mix the perturbed unlabeled target data $D_{adv}$ and the labeled target data $D_{TL}$ through \textit{Point-MixUp} \cite{ssda3d}. This step generates the adversarial mixed data $D_{AM}=\{\tilde{p}^{i}_{AM}, \hat{l}^i_{AM}\}^{N_S}_{i=1}$. In parallel, we also mix the unlabeled target data $D_{TU}$ with the labeled target data $D_{TL}$ without Adversarial Point Augmentation. This generates the point-mixed data $D_{PM}=\{\tilde{p}^{i}_{PM}, \hat{l}^i_{PM}\}^{N_S}_{i=1}$. Both mixed data $D_{AM}$ and $D_{PM}$  are used to train the student model, as described in the next section. Note that this mixup operation is conducted with the probability $P_{am}$. Without the mixup operation, we let $D_{AM}=D_{TU}$ and $D_{PM}=D_{adv}$.

\subsection{Training}

Our TODA is trained through two-stage training process. In the first stage, the teacher model $f_t$ is trained with the data  $D_{TM}$ generated by TargetMix through the detection loss, 
\begin{align}
\mathcal{L}_{det}^{tm}=\mathcal{L}^{tm}_{cls}+\mathcal{L}^{tm}_{reg}, 
\end{align}
where $\mathcal{L}_{reg}^{tm}$ and $\mathcal{L}_{cls}^{tm}$ are the standard smoothed-L1 loss and focal loss \cite{focal}, respectively. In the next stage, the student model was trained while freezing the teacher model. The loss function used to train the student model is given by 
\begin{gather}
\mathcal{L}_{adv}=\mathcal{L}^{am}_{det}+\mathcal{L}^{pm}_{det}+\lambda \mathcal{L}_{cons},
\end{gather} 
where $\lambda$ is the regularization parameter, and 
 $\mathcal{L}_{det}^{am}$ and $\mathcal{L}_{det}^{pm}$ are the detection loss terms associated with  the adversarial mixed data $D_{AM}$ and the point mixed data $D_{PM}$, respectively.  
 
 We also consider the \textit{Consistency Loss} $\mathcal{L}_{cons}$ to enforce the consistency between the detection results obtained from $D_{PM}$ and those obtained from $D_{AM}$.
Specifically, $\mathcal{L}_{cons}$ is expressed as
\begin{gather}
       \mathcal{L}^{i}_{cons}=\frac{\Sigma^{N^i_{AMB}}_{k=1}||b^{i,k}_{am}-\tilde{b}^{i,k}_{pm}||_{2}+\Sigma^{N^i_{PMB}}_{k=1}||b^{i,k}_{pm}-\tilde{b}^{i.k}_{am}||_{2}}{N^i_{AMB}+N^i_{PMB}}, \\
       \mathcal{L}_{cons}=\frac{\Sigma^{N_{TU}}_{i=1}\mathcal{L}^i_{cons}}{N_{TU}}, 
\end{gather}
where $b^{i,k}_{am}$ and $b^{i,k}_{pm}$ denote the $k$th predicted bounding boxes obtained from the $i$th sample, respectively. To quantify the distance of $b^{i,k}_{am}$ from $b^{i,k}_{pm}$, we find $\tilde{b}^{i,k}_{am}$ among the set  $\{b^{i,k}_{am}\}_{k=1}^{N_{AMB}^i}$ which is closest to $b^{i,k}_{pm}$. Similarly, we also find $\tilde{b}^{i,k}_{pm}$ among the set  $\{b^{i,k}_{pm}\}_{k=1}^{N_{PMB}^i}$ which is closest to $b^{i,k}_{am}$.
 Each bounding box is represented by center coordinates $(x,y,z)$ and size offset $(w,l,h)$ and $||.||_{2}$ denotes the $\ell_2$ norm, and $N_{AMB}^i$ and $N_{PMB}^i$ represent the number of detection boxes for the $i$th sample in $D_{AM}$ and that in $D_{PM}$, respectively.  The overall training process is presented in Algorithm \ref{alg:cap13}.

\renewcommand{\arraystretch}{1.5}
\begin{table*}[t]

  \caption{
  Comparison of domain adaptation performance on Waymo to nuScenes with different amounts of target labels for the car class. We report AP, NDS, and the closed gap based on nuScenes metric. The best adaptation result is indicated in \textbf{bold}.
  }
  \label{table:main_results}
\centering
  \begin{adjustbox}{width=1.00\textwidth}
  \begin{tabular}{c|c|c|c|c|c|c|c|c}
    \hline
    \hline
    \multirow{3}{*}{Method} & \multicolumn{2}{c|}{0.5\%} & \multicolumn{2}{c|}{1\%} & \multicolumn{2}{c|}{5\%} & \multicolumn{2}{c}{10\%} \\
    \cline{2-9} & \multirow{2}{*}{AP / NDS} & Closed Gap & \multirow{2}{*}{AP / NDS} & Closed Gap & \multirow{2}{*}{AP / NDS} & Closed Gap & \multirow{2}{*}{AP / NDS} & Closed Gap \\

 &  & \multicolumn{1}{c|}{(AP / NDS)} & & \multicolumn{1}{c|}{(AP / NDS)}  &  & \multicolumn{1}{c|}{(AP / NDS)} &  &  \multicolumn{1}{c}{(AP / NDS)}\\

    \hline
    \hline
    Source Only & 42.6 / 50.3 & +0\% / +0\% & 42.6 / 50.3 & +0\% / +0\% & 42.6 / 50.3 & +0\% / +0\% & 42.6 / 50.3 & +0\% / +0\% \\
    ST3D \cite{st3d} & 43.7 / 50.2 & +3.1\% / -0.5\% & 43.7 / 50.2 & +3.1\% / -0.5\% & 43.7 / 50.2 & +3.1\% / -0.5\% & 43.7 / 50.2 & +3.1\% / -0.5\% \\
    \hline
    Labeled Target & 36.0 / 37.7 & -18.4\% / -64.2\% & 37.2 / 38.1 & -15.1\% / -62.2\% & 61.0 / 53.2 & +51.4\% / +14.8\% & 65.6 / 58.2 & +64.2\% / +40.3\% \\
    Co-training & 47.5 / 52.7  & +13.7\% / +12.2\% & 51.4 / 54.6 & +24.6\% / +21.9\% & 57.7 / 58.0 & +42.2\% / +39.3\% & 59.4 / 58.9 & +46.9\% / +43.9\% \\
    
    SSDA3D \cite{ssda3d}& 70.3 / 65.1 & +77.4\% / +75.5\% & 73.4 / 67.1 & +86.0\% / +85.7\% & 76.2 / 68.8 & +93.9\% / +94.4\% & 78.8 / 70.9 & +101.1\% / +105.1\%  \\
    
    Ours & \bf{73.7} / \bf{67.3} & \bf{+86.9\%} / \bf{+86.7\%} & \bf{75.6} / \bf{68.5} & \bf{+92.2\%} / \bf{+92.9\%} & \bf{79.0} / \bf{71.1} & \bf{+101.7\%} / \bf{+106.1\%} & \bf{79.3} / \bf{71.4} & \bf{+102.5\%} / \bf{+107.7\%}  \\
    \hline
    Oracle & 78.4 / 69.9 & +100\% / +100\% & 78.4 / 69.9 & +100\% / +100\% & 78.4 / 69.9 & +100\% / +100\% & 78.4 / 69.9 & +100\% / +100\% \\
    \hline
    \hline
  \end{tabular}
  \end{adjustbox}

\end{table*}

\section{Experiments}

\subsection{Datasets}
We consider the setup where domain adaptation is conducted for two scenarios:
\begin{itemize}
    \item Transfer from Waymo dataset \cite{waymo} to nuScenes dataset \cite{nuscenes}
    \item Transfer from nuScenes dataset to KITTI dataset \cite{kitti}. 
    \end{itemize}
     The Waymo dataset provides 160k labeled training samples collected with 64-beam LiDAR with a 20-degree VFOV of 20 degrees (from $-17.6^\circ$ to $2.4^\circ$) while nuScenes provides 28k frames of labeled training samples collected with 32-beam LiDAR and has a VFOV of 40 degrees (from $-30.0^\circ$ to $10.0^\circ$). KITTI provides 7,481 frames of labeled training samples collected with 64-beam LiDAR and has a VFOV of 30 degrees (from $-23.6^\circ$ to $3.2^\circ$). We use 0.5\%, 1\%, 5\%, and 10\% labels for nuScenes dataset and 1\% labels for KITTI dataset, and the remaining data were utilized as unlabeled data.

We evaluated the performance of our method using both the nuScenes and KITTI 3D object detection metrics. For the nuScenes metric, we utilized the official nuScenes Detection Score (NDS) \cite{nuscenes} and Average Precision (AP) for the car category. AP represents the average precision values obtained across different thresholds of $d=0.5, 1, 2,$ and $ 4$ meters calculated based on the BEV center distance. NDS provides a comprehensive metric incorporating AP as well as errors in attributes, classification, localization, and velocity. For a fair comparison with existing methods, we also adopted the official KITTI 3D object detection metric. The KITTI metric calculates the AP for both BEV IoU and 3D IoU under an IoU threshold of 0.7 over 40 recall positions, specifically for the car category. Following the approach of ST3D \cite{st3d}, we measured the reduction in AP and NDS relative to the \textit{Oracle} performance, i.e.,  $\frac{AP_{model}-AP_{source\ only}}{AP_{oracle}-AP_{source\ only}}$ and $\frac{NDS_{model}-NDS_{source\ only}}{NDS_{oracle}-NDS_{source\ only}}$. We refer to this metric as the {\it closed gap}. The \textit{Oracle} performance was obtained by conducting supervised learning using labels from the entire target-domain dataset.
\begin{table}[t]
\renewcommand{\arraystretch}{1.2}

\caption{Comparison of domain adaptation performance on Waymo to nuScenes using SECOND-IoU based on KITTI metric. SSDA3D and TODA utilized an additional 1\% of target-domain labeled data.}
\label{table:w_n_secondiou}
\centering
\begin{tabular}{c|c|c}
\hline
\hline
Method  & $AP_{BEV}$ / $AP_{3D}$ & Closed Gap  \\ \hline \hline
Source only  & 32.9 / 17.2 & +0\% / +0\% \\  \hline
SN \cite{sn} & 33.2 / 18.6 & +1.7\% / +7.5\% \\ 
ST3D \cite{st3d}  & 35.9 / 20.2 & +15.9\% / +16.7\% \\ 
ST3D++ \cite{st3d++}  & 35.7 / 20.9 & +14.7\% / +20.9\% \\ 
L.D \cite{lidardistill}  & 40.7 / 22.9 & +41.1\% / +32.2\% \\
DTS \cite{dts}  & 41.2 / 23.0 & +43.7\% / +32.8\% \\ \hline
SSDA3D \cite{ssda3d}  & 46.6 / 29.6 & +72.1\% / +70.1\% \\ 
TODA (Ours)  & \textbf{48.1 / 30.2} & \textbf{+80.0\% / +73.4\%} \\ \hline
Oracle  & 51.9 / 34.9 & +100\% / +100\% \\
\hline
\hline
\end{tabular}

\end{table}
\subsection{Implementation Details}
We implemented the CenterPoint \cite{centerpoint} and SECOND-IoU \cite{second} using the OpenPCDet \cite{openpcdet} codebase. We followed the respective training schedule used in CenterPoint \cite{centerpoint} and SECOND-IoU. For the Waymo to nuScenes adaptation, the detection ranges for $x$, $y$, and $z$ axes were set to $[-54.0, 54.0]$, $[-54.0, 54.0]$, and $[-5.0, 4.8]$ meters, and the voxel size was set to $[0.075, 0.075, 0.2]$. In the case of nuScenes to KITTI adaptation, the detection ranges in $x$, $y$, and $z$ axes were set to $[-76.2, 76.2]$, $[-76.2, 76.2]$, and $[-3.0, 5.0]$ meters with the same voxel size of $[0.075, 0.075, 0.2]$. The intensity of the point cloud was normalized to have a value between $[0,1]$. We applied data augmentation techniques, including random flip along X and Y axes, random rotation, random scaling, and GT sampling. GT sampling was only applied to the labeled data in the target domain for the second stage. The probability $P_{tm}$ used in TargetMix was set to 0.4. The probability $P_{am}$ in AdvMix was set to 0.6. The hyperparameters $\lambda$, $\rho$, and $\epsilon$ were empirically chosen to 1, 0.5, and 0.001, respectively. All experiments were conducted on four 24GB RTX 3090TI GPUs.



\begin{table}[t]
\renewcommand{\arraystretch}{1.2}

    \caption{Comparison of domain adaptation performance on nuScenes to KITTI using SECOND-IoU based on KITTI metric. SSDA3D and TODA utilized an additional 1\% of target-domain labeled data.}
        \label{table:n_k_secondiou}
    \centering
    \begin{tabular}{c|c|c}
    \hline
    \hline
        Method  & $AP_{BEV}$ / $AP_{3D}$ & Closed Gap  \\ \hline \hline
        Source only & 51.8 / 17.9 & +0\% / +0\% \\ \hline
        SN \cite{sn} & 59.7 / 37.6 & +25.1\% / +35.4\% \\ 
        ST3D \cite{st3d} & 75.9 / 54.1 & +76.6\% / +59.5\% \\
        ST3D++ \cite{st3d++} & 80.5 / 62.4 & +91.1\% / +80.0\% \\
        DTS \cite{dts} & 81.4 / 66.6 & +94.0\% / +87.6\% \\ \hline
        SSDA3D \cite{ssda3d} & 81.5 / 67.4 & +94.3\% / +89.0\% \\
        TODA (Ours) & \textbf{82.7 / 68.6} & \textbf{+98.1\% / +91.2\%} \\ \hline
        Oracle & 83.3 / 73.5 & +100\% / +100\% \\
        \hline
\hline
    \end{tabular}

\end{table}

\begin{table}[t]
\renewcommand{\arraystretch}{1.2}
\caption{Ablation study for evaluating the contribution of each component of TODA on the nuScenes validation dataset.}
  \label{table:ablation_main}
\centering
  \begin{adjustbox}{width=0.4\textwidth}
  \begin{tabular}{c|c|c|c}
    \hline
    \hline
     Method &  TargetMix  & AdvMix & AP / NDS \\
    \hline
    \hline
    Source Only & &  & 37.2 / 38.1\\
    Co-training & &  & 51.4 / 54.6 \\ 
    \hline
    \multirow{2}{*}{TODA} & $\checkmark$ &  & 71.2 / 65.9\\
     & $\checkmark$ & $\checkmark$ & 75.6 / 68.5 \\
    \hline
    \hline
  \end{tabular}
  \end{adjustbox}

\end{table}
\subsection{Main Results}

We evaluated the performance of our SSDA model using CenterPoint. In the nuScenes dataset, we utilized labels for 0.5\%, 1\%, 5\%, and 10\% of the target domain data, corresponding to 141, 282, 1407, and 2813 frames, respectively. Table \ref{table:main_results} presents the performance of TODA in the Waymo to nuScenes adaptation task, using the nuScenes metric. We compare TODA with the existing domain adaptation methods including {\it Co-training}, {\it Labeled Target}, ST3D \cite{st3d}, and SSDA3D \cite{ssda3d}. \textit{Co-training} utilized supervised learning using the labeled data in both source and target domains  while \textit{Labeled Target} employed supervised learning solely utilizing the labeled data in the target domain. Table \ref{table:main_results} shows that TODA consistently outperforms all other methods by significant margins. When 0.5\% of labeled data are used in the target domain, TODA demonstrates notable performance gains of 3.4\% in AP and 2.2\% in NDS over SSDA3D, the current state-of-the-art method. TODA also achieves more than 85\% closed gap from the Oracle performance. Even with  1\% labeled target-domain data, TODA achieves improvements of 2.2\% in AP and 1.4\% in NDS over SSDA3D. It also achieves above 90\% closed gap in both metrics. Remarkably, when 5\% and 10\% labeled  data are utilized, TODA even surpasses the Oracle performance. This would be possibly because TODA leverages both source-domain and target-domain data, while the Oracle performance is obtained using the target-domain dataset only. In summary, TODA achieves performance comparable to the Oracle performance while significantly reducing annotation costs.

\begin{table}[t]
\renewcommand{\arraystretch}{1.2}
\caption{Ablation study for the TargetMix module.  Polar, Enhance., and Match., indicate Polar Coordinate-based Mix, Enhanced Mix Strategy, and LiDAR Distribution Matching, respectively.}
  \label{table:targetmix_ablation}
\centering
  \begin{adjustbox}{width=0.4\textwidth}
  \begin{tabular}{c|c|c|c|c}
    \hline
    \hline
    Method & Polar & Enhance. & Match. & AP / NDS\\
    \hline
    \hline
    CutMix \cite{ssda3d} & & & & 66.8 / 63.4 \\
    \hline
    \multirow{3}{*}{TargetMix} & $\checkmark$ & & & 68.3 / 63.8 \\
    & $\checkmark$ & $\checkmark$ &   & 69.6 / 64.5 \\  
    & $\checkmark$ & $\checkmark$ & $\checkmark$ & 71.2 / 65.9 \\
    \hline
    \hline
  \end{tabular}
  \end{adjustbox}

\end{table}
\renewcommand{\arraystretch}{1}
\begin{table}[t]
\renewcommand{\arraystretch}{1.2}
  \caption{Ablation study for the AdvMix module. Adv., Cons., MixUp., indicate Adversarial Point Augmentation, Consistency Loss, and Point-MixUp, respectively.}
  \label{table:advmix_ablation}
\centering
  \begin{adjustbox}{width=0.8\columnwidth}
  \begin{tabular}{c|c|c|c|c}
    \hline
    \hline
    Method & Adv.  & Cons. & MixUp  & AP / NDS\\
    \hline
    \hline
    TargetMix & & & & 71.2 / 65.9 \\
    \hline
    \multirow{3}{*}{AdvMix} & $\checkmark$ & & & 73.2 / 67.1 \\
     & $\checkmark$ & $\checkmark$ & & 73.5 / 67.2 \\ 
     & $\checkmark$ & $\checkmark$ & $\checkmark$ & 75.6 / 68.5\\

    \hline
    \hline
  \end{tabular}
  \end{adjustbox}

\end{table}




\begin{table}[t!]
    \caption{Ablation study for the probability of mixing augmentation $P_{tm}$ and $P_{am}$.}
    \centering
    \subfloat[]{
        \begin{tabular}{c|c}
        \hline \hline
        $P_{tm}$ & AP \\ \hline \hline
        0.1 & 70.1 \\ \hline
        \rowcolor{gray} 0.2 & \textbf{71.2} \\ \hline
        0.3 & 70.1 \\ \hline
        0.4 & 69.8 \\ \hline \hline
        \end{tabular}
        \label{tabl:abl_ptm}
    }
   \hspace{0.5cm}
    \subfloat[]{
        \begin{tabular}{c|c}
        \hline \hline
        $P_{am}$ & AP \\ \hline \hline
        0.4 & 78.1 \\ \hline
        0.5 & 78.5 \\ \hline
        \rowcolor{gray} 0.6 & \textbf{79.0} \\ \hline
        0.7 & 78.8 \\ \hline \hline
        \end{tabular}
        \label{tabl:abl_pam}
    }
    \label{table:abl_pam_ptm}
\end{table}
\begin{figure}[t]
    \centering
    \includegraphics[width=0.8\columnwidth]{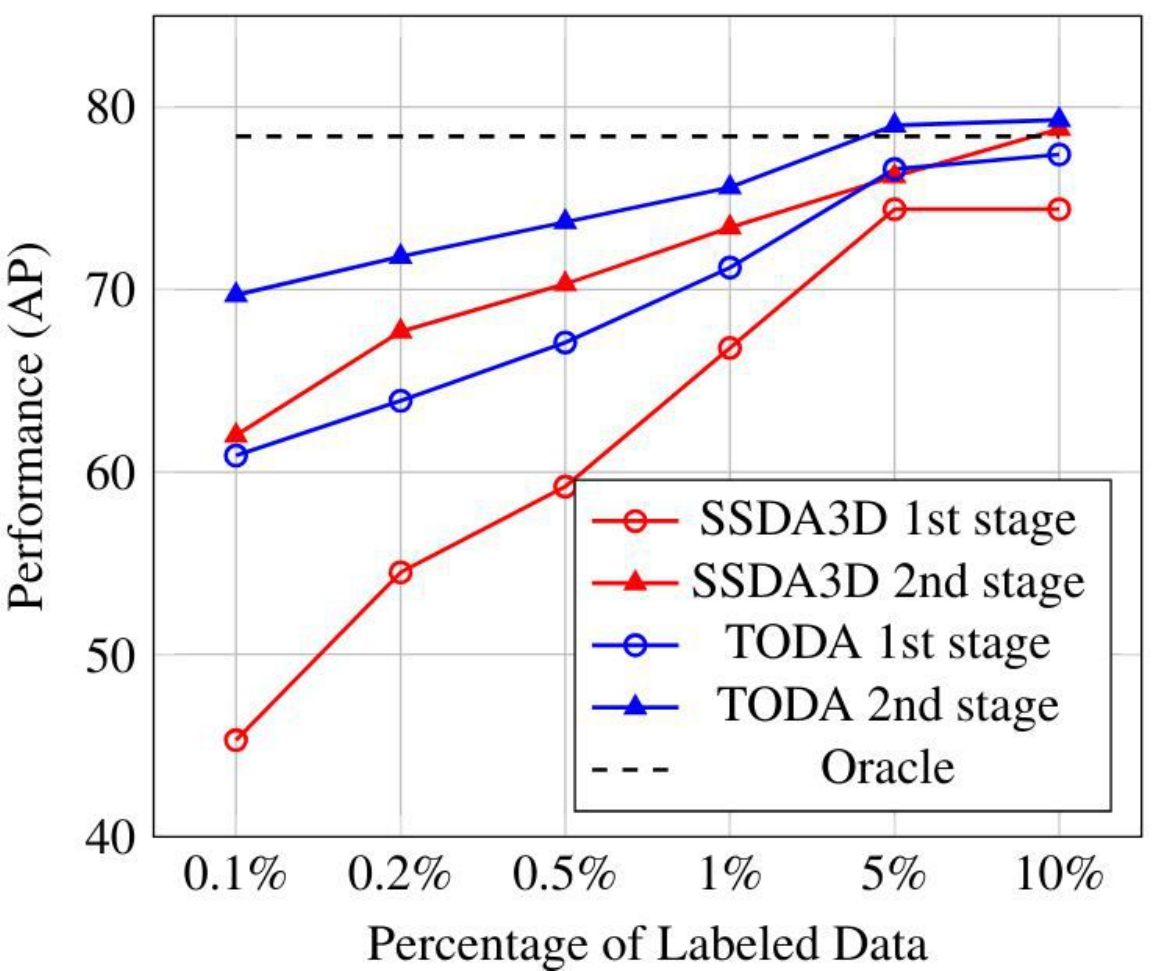}
    \caption{{\bf Comparison with SSDA3D in each stage for different percentages:} Performance comparison between TODA and SSDA3D across various sizes of labeled data (0.1\%, 0.2\%, 0.5\%, 1\%, 5\%, and 10\%)}
    \label{fig:toda_each_stage}
\end{figure}

\begin{table}[t]
\renewcommand{\arraystretch}{1.2}
\caption{The performance of TODA with 0.1\% and 0.2\% of target-domain labeled data.}
\centering
\begin{adjustbox}{width=0.5\columnwidth}
\begin{tabular}{c|c|c}
\hline
\hline
{Method} & 0.1\% & 0.2\% \\
\hline
\hline
Labeled Target & fail & fail  \\
Co-training & 50.3  & 52.2  \\ 
\hline
SSDA3D \cite{ssda3d}& 62.0 & 67.6 \\ 
TODA & \textbf{69.7} & \textbf{71.8}  \\  
\hline
Oracle & 78.4 &  78.4  \\
\hline
\hline
\end{tabular}
\end{adjustbox}
\vspace{-0.4cm}
\label{table:main}
\end{table}


We further validate the effectiveness of our method in comparison with various latest UDA techniques including SN \cite{sn}, ST3D \cite{st3d}, ST3D++ \cite{st3d++}, L.D \cite{lidardistill}, and DTS \cite{dts}. For fair comparison, we implemented TODA on the SECOND-IoU model and conducted evaluation using the KITTI metrics for the Waymo to nuScenes adaptation task. Table \ref{table:w_n_secondiou} shows that the proposed TODA method still maintains performance gains over other UDA methods. While UDA methods exhibit limitation in reducing the domain gap, TODA achieves a significantly higher closed gap of +80.0\% using only 1\% of the labeled target data.

Table \ref{table:n_k_secondiou} shows the performance of TODA on the nuScenes to KITTI adaptation scenario as well. Although the UDA methods achieve performance close to the oracle, TODA generalizes well to the nuScenes to KITTI setup. Notably, TODA achieves a closed gap of 98.1/91.2\% relative to the Oracle. These results highlight the effectiveness of our SSDA approach in adapting to different target domains, even with 1\% of labeled data.

\subsection{Ablation Studies}
\subsubsection{Contribution of Each Component } 
We conducted ablation studies to assess the contribution of each component of TODA to the overall performance. The evaluation was conducted using the nuScenes validation set on the Waymo to nuScenes task. Table~\ref{table:ablation_main} presents the results when  1\% of the target-domain labeled data are used. We sequentially enabled \textit{TargetMix} and \textit{AdvMix} on top of the baseline method based on Co-training. The inclusion of \textit{TargetMix} results in a notable performance gain of 19.8\% in AP and 11.3\% in NDS, highlighting its significant contribution. Incorporating \textit{AdvMix} leads to a performance improvement of 4.4\% in AP and 2.6\% in NDS.



\begin{figure*}[t]
    \subfloat[][SSDA3D]{\includegraphics[width=0.5\textwidth]{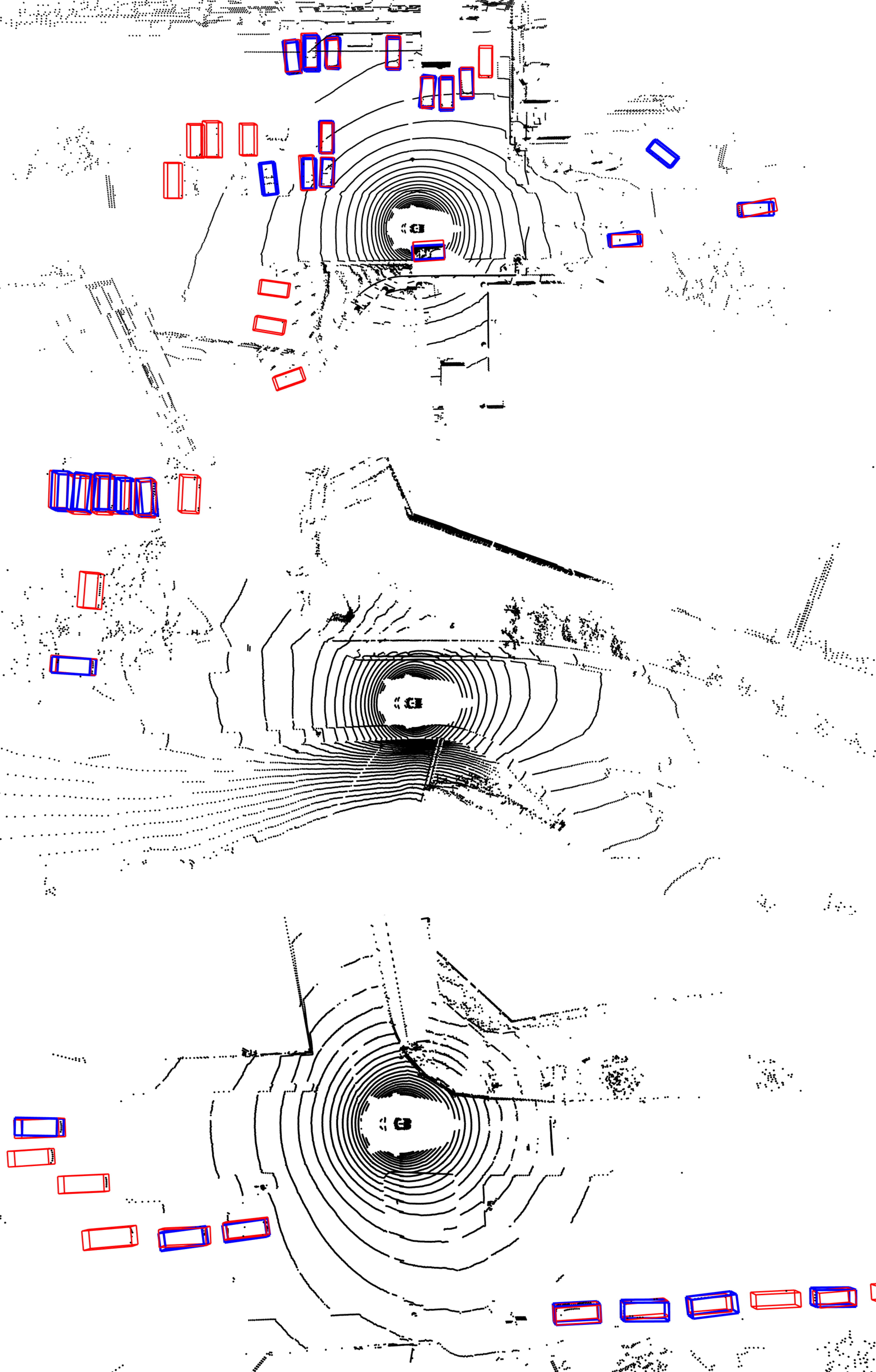}}
    \subfloat[][Ours]{\includegraphics[width=0.5\textwidth]{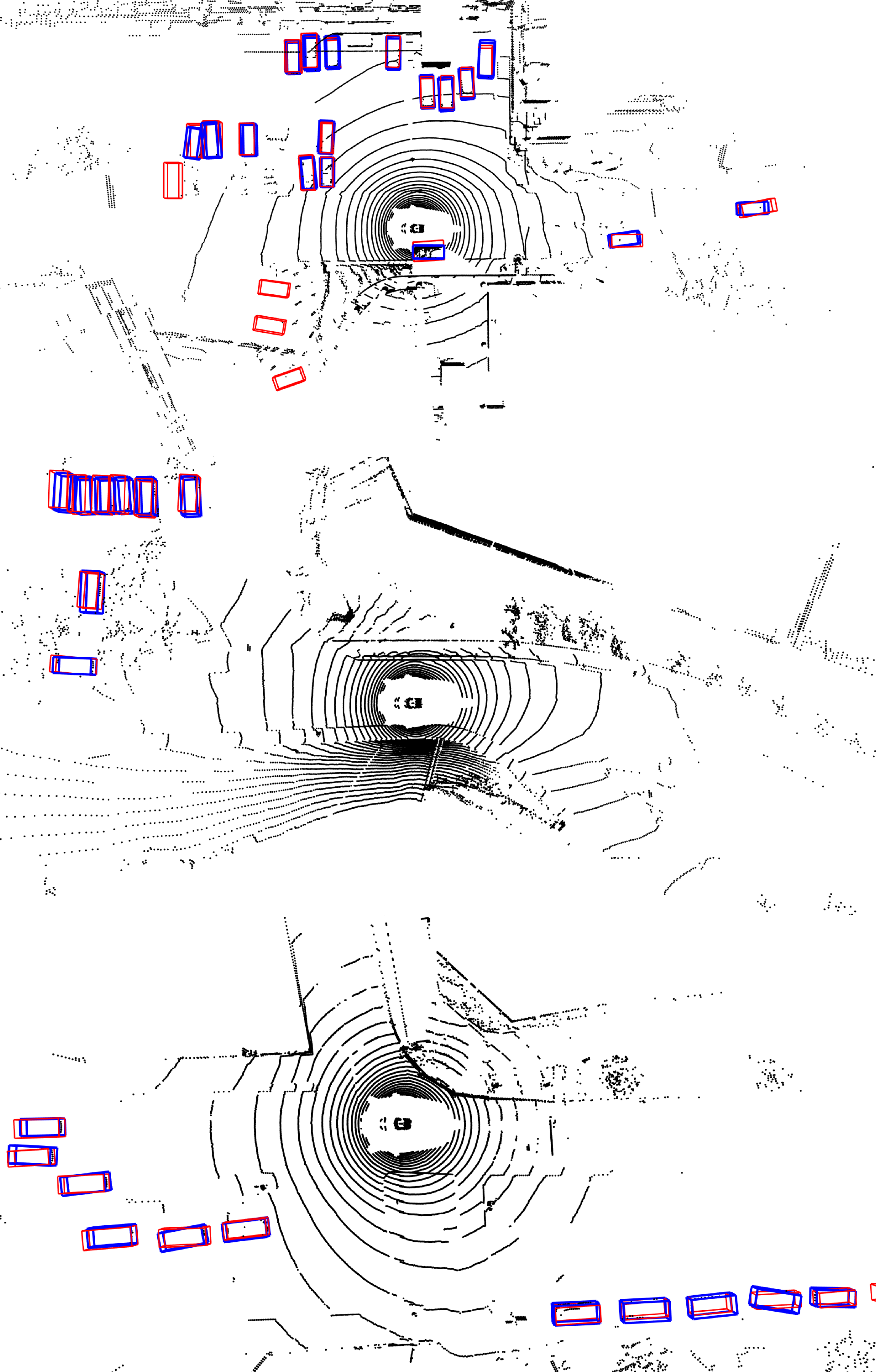}}
    \caption{{\bf Comparison of detection results:}  (a) SSDA3D and (b) TODA. All samples are from nuScenes val split. The red box represents the ground truth, and the blue box indicates the predicted bounding box.}
    \label{fig:qualitative}
\end{figure*}

\subsubsection{Sub-components of TargetMix}  Table \ref{table:targetmix_ablation} illustrates the ablation study evaluating the contribution of three components within the TargetMix module: \textit{Polar Coordinate-based Mix}, \textit{Enhanced Mix Strategy}, and \textit{LiDAR Distribution Matching}. The \textit{Enhanced Mix Strategy} refers to the idea of segmenting the area into $K$ segments and allocating the LiDAR points accordingly. LiDAR Distribution Matching refers to the method that adjusts LiDAR data to match the configuration of the target-domain LiDAR. While the naive \textit{Polar Coordinate-based Mix} demonstrates only a slight improvement over the CutMix baseline, its impact becomes significant when combined with our \textit{Enhanced Mix Strategy}. This {\it Enhanced Mix Strategy} achieves a performance gain of 2.8\% in AP and 2.1\% in NDA. Additionally, the \textit{LiDAR Distribution Matching} method further enhances performance by 1.6\% in AP and 1.4\% in NDS.


\subsubsection{Sub-components of AdvMix} Table \ref{table:advmix_ablation} presents the ablation study assessing the impact of each component within the AdvMix module. We utilized a model trained with TargetMix as our baseline. Three components were considered: \textit{Adversarial Point Augmentation}, \textit{Consistency Loss}, and \textit{Point-Mixup}. \textit{Adversarial Point Augmentation} results in a performance gain of 2.0\% in AP and 1.2\% in NDS. Additionally, \textit{Consistency Loss} contributes to an increase in AP by 0.3\% and NDS by 0.1\%. Finally, \textit{Point-MixUp} leads to further improvements of 2.1\% in AP and 1.3\% in NDS.

\subsubsection{Probabilities $P_{tm}$ and $P_{am}$ of TargetMix and AdvMix} Table \ref{table:abl_pam_ptm} presents the performance of TODA as a function of the parameters $P_{tm}$ and $P_{am}$. Table \ref{table:abl_pam_ptm} (a) shows the performance of the first stage with varying $P_{tm}$ values, while Table \ref{table:abl_pam_ptm} (b) illustrates the final TODA performance with varying $P_{am}$ values using a model trained with $P_{tm} = 0.2$. The results demonstrate that TODA exhibits robust performance with respect to both $P_{tm}$ and $P_{am}$, with only slight degradation under different settings. The optimal performance is achieved with $P_{tm} = 0.2$ and $P_{am} = 0.6$.

\subsubsection{Scenarios Using Extremely Low Number of Labels} In Table \ref{table:main}, we evaluate the performance of TODA in scenarios where the percentage of labeled target-domain data is extremely low, e.g.,  0.1\% (28 frames) and 0.2\% (56 frames).  Training exclusively with such small number of labeled targets failed due to insufficient target-domain data. In contrast, TODA achieves AP values of 69.7\% and 71.8\% respectively, surpassing SSDA3D by 6.3\% and 4.2\% respectively. Remarkably, TODA achieves 75\% and 81\% of closed gap even with  0.1\% and 0.2\% labeled data respectively. These results demonstrate the potential of TODA in highly data-constrained environments.

Figure \ref{fig:toda_each_stage} presents a performance comparison between TODA and SSDA3D across various sizes of labeled data (0.1\%, 0.2\%, 0.5\%, 1\%, 5\%, and 10\%). Performance was assessed at both the initial and secondary stages of each detector. We noted an increasing performance gap between TODA and SSDA3D at both stages as the percentage of labeled data decreased. This trend seems to be attributed to proposed LiDAR Distribution Matching that conducts the  adaptation of LiDAR data distribution without using labeled data. This leads to   notable performance improvements for TODA in the initial stage. Additionally, Adversarial Point Augmentation reshapes the distribution of unlabeled data in the target domain, enabling more effective feature alignment between labeled and unlabeled data. This boosts the performance of our pseudo-label based semi-supervised learning in the second stage.
\subsection{Qualitative Results}
We present some qualitative results. We considered the setup where 1\% of the labeled data are used in the target domain. The experiments were conducted on the nuScenes validation set.

In Fig. \ref{fig:tsne}, we present visualizations of feature distributions utilizing t-SNE \cite{tsne} under both pre- and post-application of Adversarial Point Augmentation. Fig. \ref{fig:tsne} (a) highlights the distribution gap between the labeled and unlabeled data, while Fig. \ref{fig:tsne} (b) shows the gap  between  the labeled  data and the adversarially perturbed unlabeled data. Notably, Adversarial Point Augmentation demonstrates its effectiveness in mitigating the distribution shift.



Fig. \ref{fig:qualitative} illustrates the detection results produced by both the existing method and the proposed TODA method.  The figures in the left column depict the detection results attained by SSDA3D, while those in the right column show the results of the  proposed TODA method. We observe that TODA demonstrates enhanced detection results, effectively identifying objects that were previously either missed or inaccurately detected by SSDA3D.

\section{Conclusions}

In this paper, we introduced TODA, an SSDA framework for 3D object detection based on a target-oriented domain augmentation strategy. To mitigate the disparity in data distribution between the source and target domains, we introduced TargetMix. TargetMix utilizes an inter-domain mixup augmentation strategy within a polar coordinate system, considering the LiDAR scanning mechanism. Additionally, TargetMix incorporates LiDAR Distribution Matching to adapt the source domain data to align with the configurations of the target-domain LiDAR sensor. Additionally, we proposed AdvMix, which adds adversarial perturbation to the unlabeled data to mitigate intra-domain disparity. We optimized the perturbation direction to maximize detection performance, enabling AdvMix to generate consistent representations of both labeled and unlabeled data in the target domain.
By integrating TargetMix and AdvMix, TODA effectively utilizes both labeled and unlabeled data for domain adaptation. Our evaluation demonstrated that TODA achieved significant performance gains over existing domain adaptation methods and approached performance levels close to the Oracle performance.

Moving forward, we aim to enhance TODA's capabilities to cope with adverse weather conditions and low-resolution LiDAR environments, as well as explore its potential for other sparse data modalities like radar. These advancements will expand TODA's usefulness and offer promising avenues for future research in domain adaptation and 3D object detection.

\section*{Acknowledgments}
This work was partly supported by Institute of Information \& communications Technology Planning \& Evaluation (IITP) grant funded by the Korea government (MSIT) [No.2021-0-01343-004, Artificial Intelligence Graduate School Program (Seoul National University)], the National Research Foundation of Korea (NRF) grant funded by the Korea government (MSIT) [No.2020R1A2C2012146], and Information \& communications Technology Promotion(IITP) grant funded by the Korea government(MSIP) [No.2021-0-01314, Development of driving environment data stitching technology to provide data on shaded areas for autonomous vehicles].

\bibliographystyle{IEEEtran}
\bibliography{IEEEtran}


\section*{Biography Section}
\vspace{-0.8cm}
\begin{IEEEbiography}[{\includegraphics[width=1in,height=1.25in,clip,keepaspectratio]{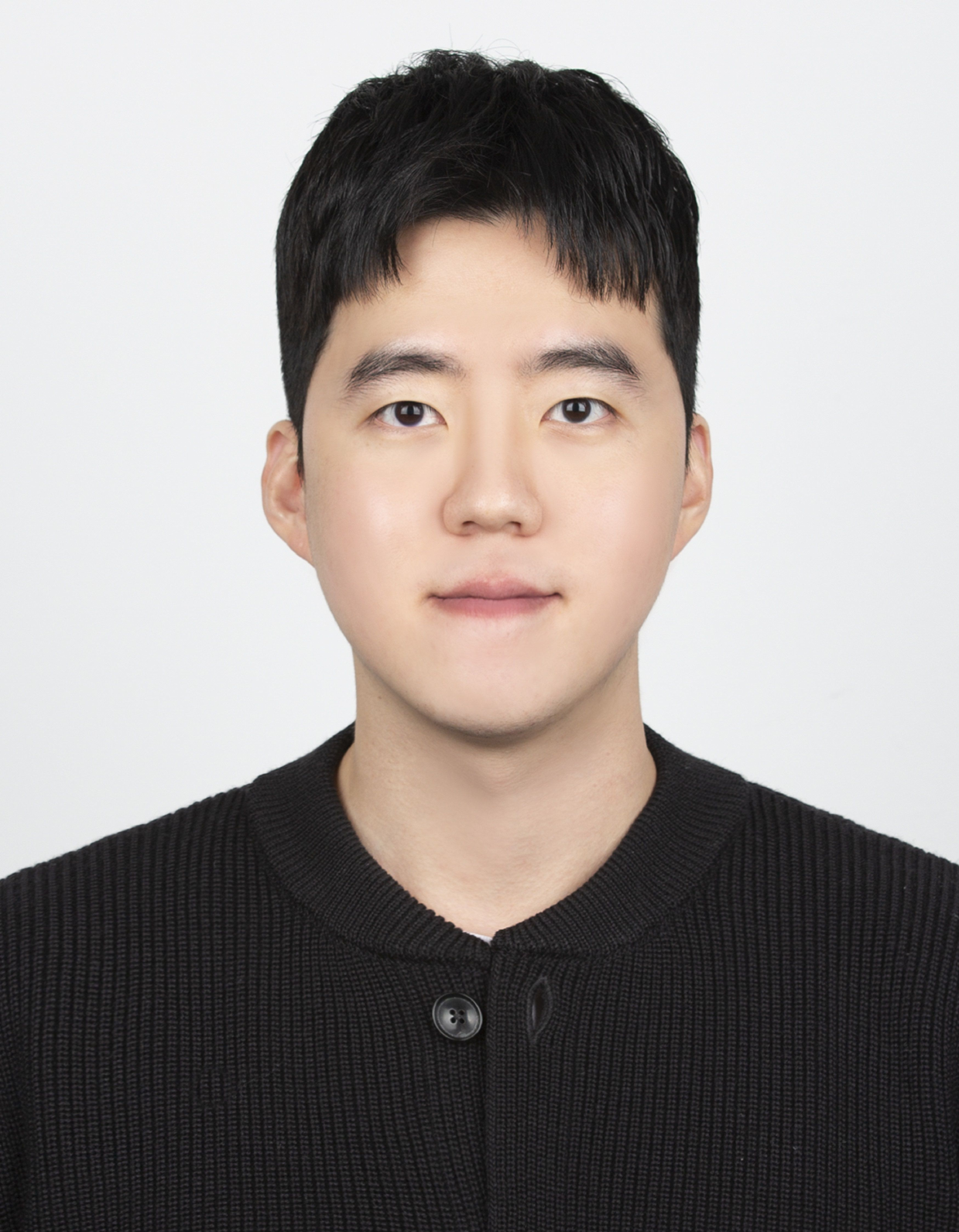}}]{Yecheol Kim} received a B.S. degree in Electrical Engineering from the Hanyang University, Seoul, South Korea, in 2018. He is currently pursuing a Ph.D. degree in the Hanyang University. His research interests include multi-modal object detection, domain adaptation, and efficient architecture for autonomous driving.
\end{IEEEbiography}
\vspace{-0.8cm}
\begin{IEEEbiography}[{\includegraphics[width=1in,height=1.25in,clip,keepaspectratio]{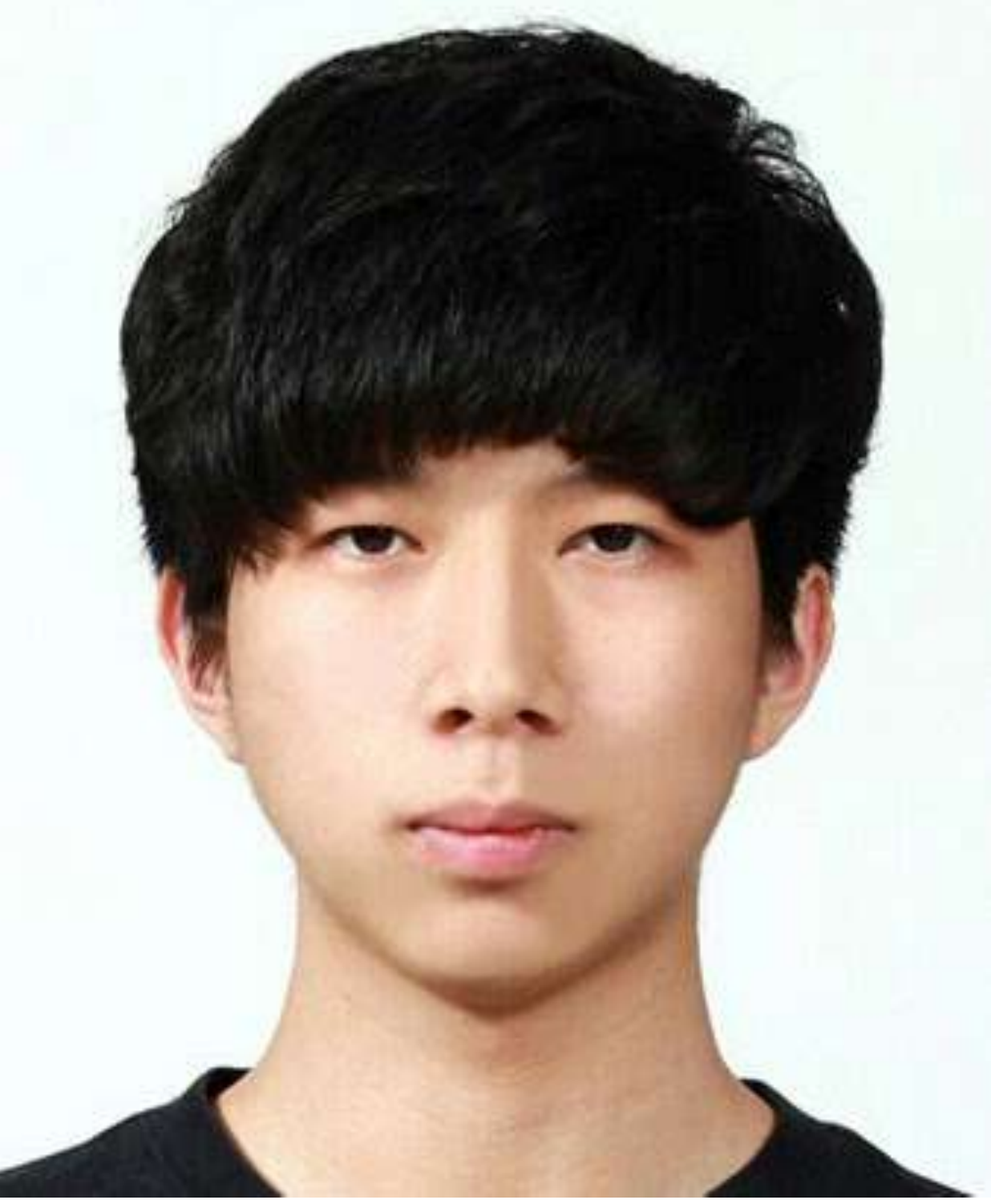}}]{Junho Lee} received a B.S. degree in Electrical Engineering from the Hanyang University, Seoul, South Korea, in 2018. He is currently pursuing a Ph.D. degree in the Hanyang University. His research interests reinforcement learning, domain adaptation, and domain generalization for autonomous driving.
\end{IEEEbiography}
\vspace{-0.8cm}
\begin{IEEEbiography}[{\includegraphics[width=1in,height=1.25in,clip,keepaspectratio]{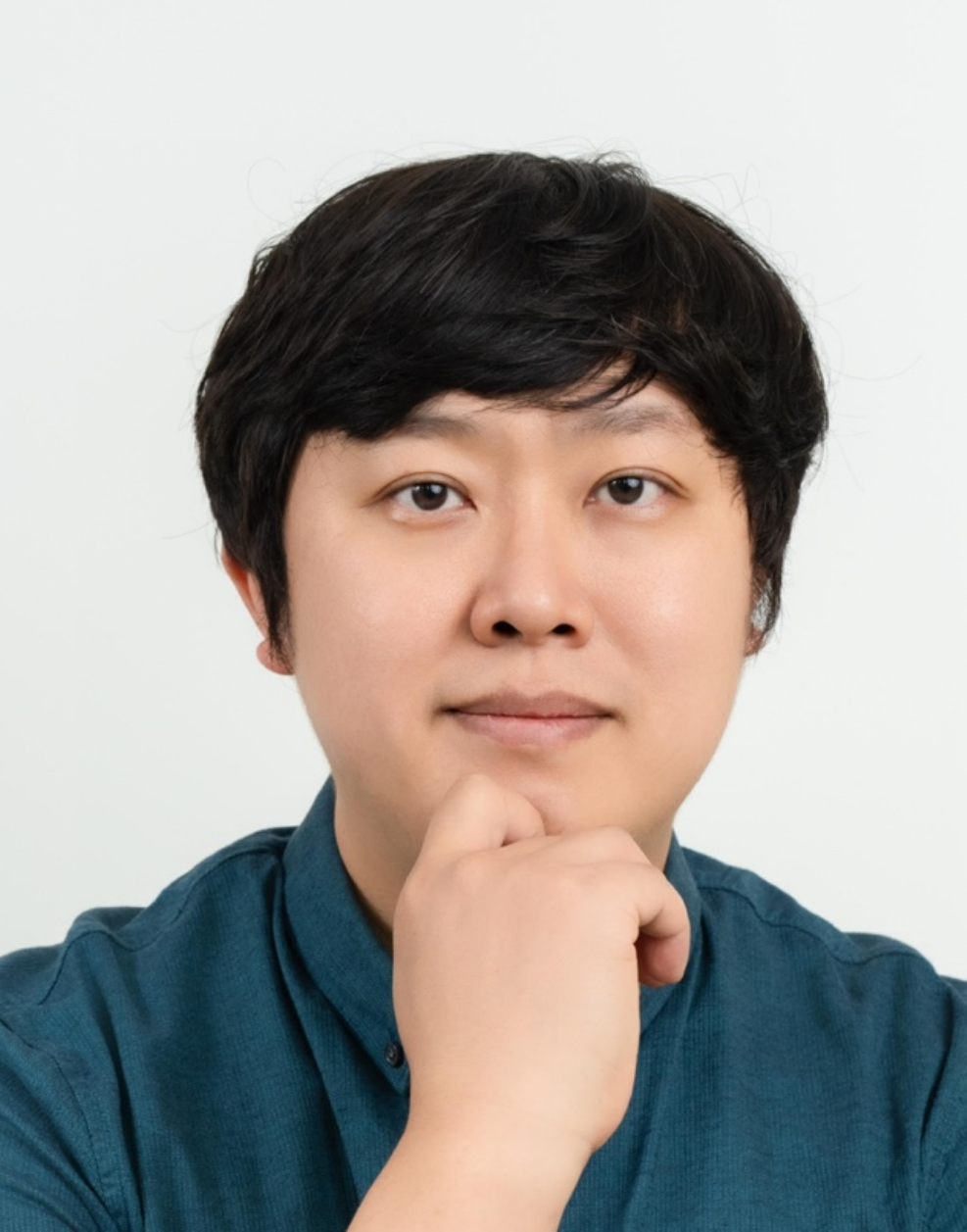}}]{Changsoo Park} received B.S, M.S and Ph.D. degrees from the Electrical Engineering, KAIST. In 2015, he joined Samsung Electronics, where he participated in research on autonomous driving and robot. Since 2019, he has been researching autonomous driving at Kakao mobility.
\end{IEEEbiography}
\vspace{-0.8cm}

\begin{IEEEbiography}[{\includegraphics[width=1in,height=1.25in,clip,keepaspectratio]{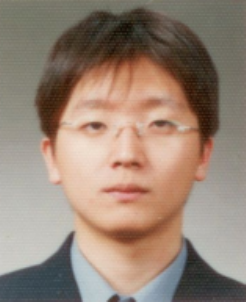}}]{Hyung Won Kim} received the B.S., M.S. and Ph.D. degrees in Electronic Engineering from Hanyang University.
From 2015 to 2021, he was with the Hyundai Mobis where his major research topics included target tracking and sensor fusion.
He has been working at Kakao Mobility since 2021. His research area includes sensor fusion and machine learning.
\end{IEEEbiography}
\vspace{-0.8cm}
\begin{IEEEbiography}[{\includegraphics[width=1in,height=1.25in,clip,keepaspectratio]{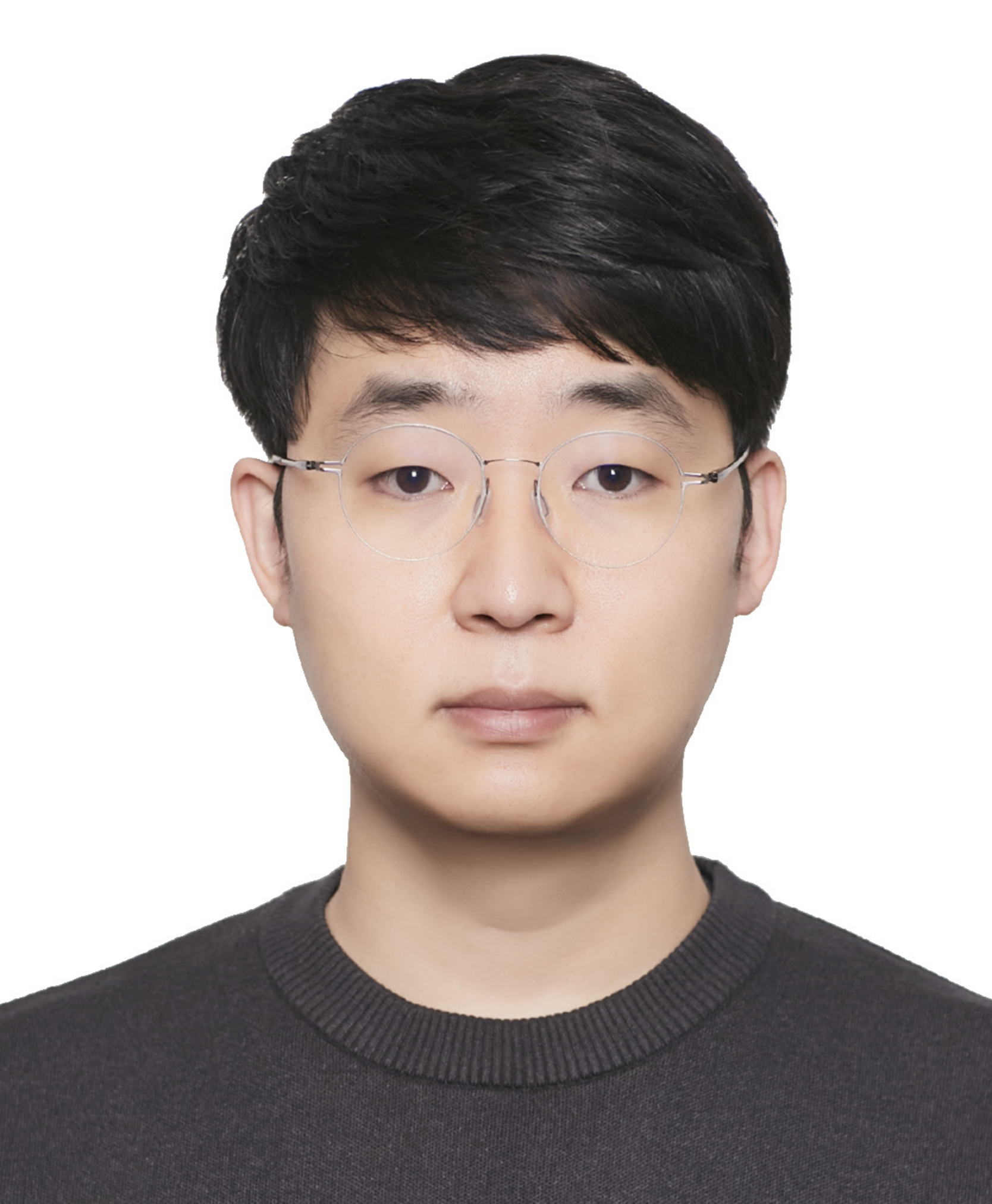}}]{Inho Lim} received B.S degres from the Information and Technology Department, Ajou University. In 2012, he joined Samsung Electronics, Suwoun, Korea, where he participated in the software engineering research team. Since 2017, he has participated in autonomous driving SW research at Samsung Advanced Institute of Technology. Since 2019, he has been researching the Localization and Perception technologies for autonomous driving in Kakao Mobility Corp.
\end{IEEEbiography}
\vspace{-0.8cm}
\begin{IEEEbiography}[{\includegraphics[width=1in,height=1.25in,clip,keepaspectratio]{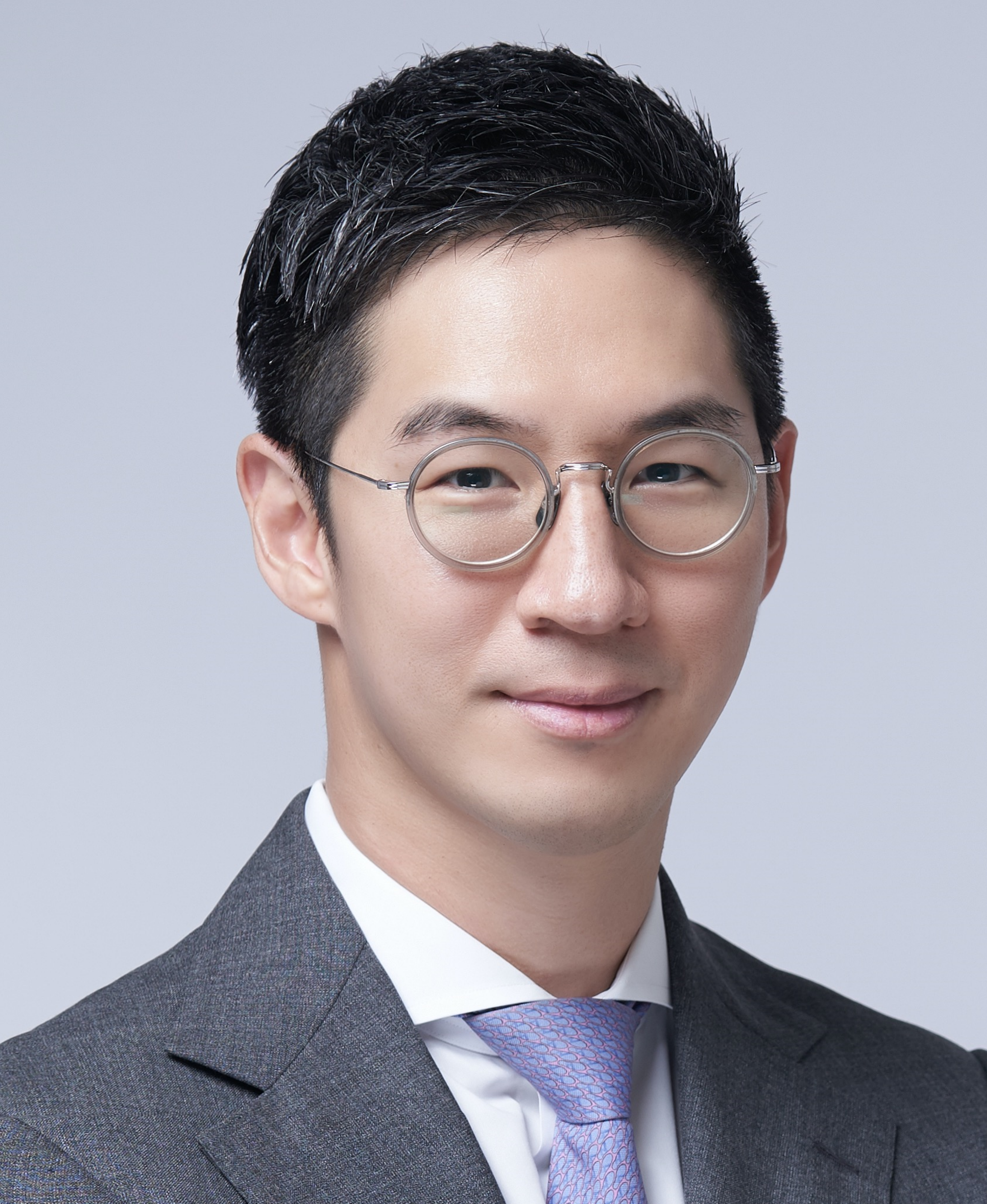}}]{Christopher Chang} Christopher Chang received a B.S. degree from Department of Electrical Engineering, Seoul National University, and M.S. and Ph.D. degrees in Electrical Engineering from California Institute of Technology, Pasadena CA. Since 2012, he had been working with Samsung Electronics and Hyundai Motor Company, where he developed corporate strategies and business roadmaps. In 2020, he joined Kakao Mobility Corp. where he has been leading strategy, business development, and R\&D for the next generation mobilities including autonomous driving, robotics, digital twin, and urban air mobility.
\end{IEEEbiography}
\vspace{-0.8cm}
\begin{IEEEbiography}[{\includegraphics[width=1in,height=1.25in,clip,keepaspectratio]{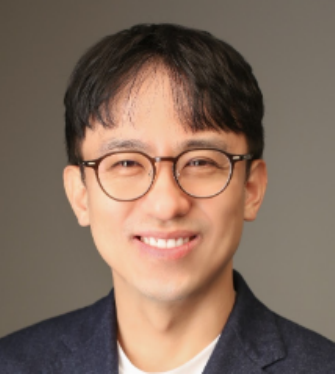}}]{Jun Won Choi} earned his B.S. and M.S. degrees from Seoul National University and his Ph.D. from the University of Illinois at Urbana-Champaign. Following his studies, he joined Qualcomm in San Diego, USA, in 2010. From 2013 to 2024, he served as a faculty member in the Department of Electrical Engineering at Hanyang University. Since 2024, he has held a faculty position in the Department of Electrical and Computer Engineering at Seoul National University. He currently serves as an Associate Editor for both IEEE Transactions on Intelligent Transportation Systems, IEEE Transactions on Vehicular Technology, International Journal of Automotive Technology. His research spans diverse areas including signal processing, machine learning, robot perception, autonomous driving, and intelligent vehicles.
\end{IEEEbiography}

\vfill

\end{document}